\documentclass[11pt,a4paper]{article}
\usepackage[table]{xcolor}
\usepackage[hyperref]{eacl2021}
\usepackage{times}
\usepackage{latexsym}
\usepackage{graphicx}

\usepackage{todonotes}
\usepackage{amsmath}
\usepackage{booktabs}
\usepackage{float}

\usepackage{enumitem}
\usepackage{amsmath}
\usepackage{amssymb}
\usepackage{caption}
\usepackage{subcaption}
\usepackage{titlesec}
\restylefloat{table}
\DeclareMathOperator*{\argmax}{argmax}
\usepackage{microtype}

\aclfinalcopy 

\title{Unsupervised Machine Translation On Dravidian Languages}
 \author{
        Sai Koneru, Danni Liu and Jan Niehues\\
        Department of Data Science and Knowledge Engineering, Maastricht University \\ 
        \texttt{s.koneru@student.maastrichtuniversity.nl}\\
        \texttt{\{danni.liu,jan.niehues\}@maastrichtuniversity.nl}
        }

\date{}
\begin{document}
\maketitle
\begin{abstract}
Unsupervised neural machine translation (UNMT) is beneficial especially for low resource languages such as those from the Dravidian family. 
However, UNMT systems tend to fail in realistic scenarios involving actual low resource languages.
Recent works propose to utilize auxiliary parallel data and have achieved state-of-the-art results.
In this work, we focus on unsupervised translation between English and Kannada, a low resource Dravidian language.
We additionally utilize a limited amount of auxiliary data between English and other related Dravidian languages. 
We show that unifying the writing systems is essential in unsupervised translation between the Dravidian languages.
We explore several model architectures that use the auxiliary data in order to maximize knowledge sharing and enable UNMT for distant language pairs. 
Our experiments demonstrate that it is crucial to include auxiliary languages that are similar to our focal language, Kannada.
Furthermore, we propose a metric to measure language similarity and show that it serves as a good indicator for selecting the auxiliary languages.
\end{abstract}
\section{Introduction}

Around 200 million people in the world speak a Dravidian language which has almost 80 varieties \citep{kolipakam2018bayesian}. 
Given the significant number of speakers, improving translation quality on these languages is an important task to facilitate intercultural communication. 
While neural machine translation (NMT) systems \citep{bahdanau-etal-2015-neural,vaswani2017attention} led to great improvements on high resource languages, a major challenge remains when there is little parallel data between the source and target languages. 
Compared to resource-rich languages like German and French, the amount of high-quality parallel data for Dravidian languages is minimal. 
As a result, the current translation quality for Dravidian languages is largely lagging behind.

Unsupervised neural machine translation (UNMT;  \citeauthor{artetxe2018unsupervised}, \citeyear{artetxe2018unsupervised}, \citeauthor{lample2018unsupervised}, \citeyear{lample2018unsupervised}) learns to translate between languages without relying on parallel data.
Instead, only monolingual corpora are required.
This relieves the need to acquire large parallel datasets, which is especially difficult for low resource languages.
Despite this attractive property, existing literature \citep{marchisio-etal-2020-unsupervised,kim-etal-2020-unsupervised} suggests that UNMT performs well under the following conditions: 1) Source and target languages are similar; 2) Monolingual data for both the languages belong to the same domain.
In most realistic scenarios that call for unsupervised translation, these conditions are rarely fulfilled.

Considering these challenges, one promising research direction to address these conditions is to combine multilingual NMT models with unsupervised translation \citep{liu2020mbart,li-etal-2020-reference,garcia2020harnessing,guzman-etal-2019-flores}. 
Incorporating similar high resource languages could alleviate the data scarcity of low resource languages by exploiting the language similarity between them. 
\citet{fraser-2020-findings} describes this scenario as ``unsupervised machine translation with multilingual transfer''. 
For example, \citet{li-etal-2020-sjtu} investigate translating between German to Upper Sorbian with no parallel data between them but using parallel data between the high resource languages German and English. 
Among Dravidian languages, the comparatively better resourced ones are Tamil, Telugu, Malayalam and Kannada.
In turn, among these four languages, Kannada is relatively low resource compared to the others
\cite{reddy-sharoff-2011-cross}.
Therefore, we consider a realistic scenario of improving translation quality on Kannada using auxiliary data in the other three languages.
We refer to the additional language as \textit{reference language} following the terminology in \citet{li-etal-2020-reference}.

\begin{figure}[!ht]
\includegraphics[width=0.4\textwidth]{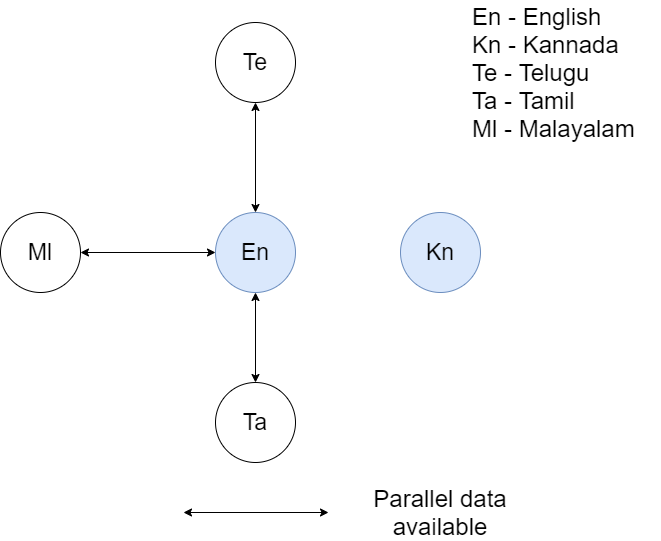}
\caption{Illustration of the data conditions of the languages in our work. 
Connected nodes indicate availability of parallel data. 
We focus on translating between English and Kannada (blue nodes) without any parallel data.}
\label{fig:setup}
\end{figure}

In this work, with a focus on Dravidian languages, we analyze the influence of related languages as auxiliary data in various setups of multilingual UNMT (MUNMT). 
As shown in Figure 1, we focus on a scenario of translating between Kannada and English with the following conditions: 1) no parallel data for Kannada-English; 2) very limited in-domain parallel data  ($\sim$30k sentences) between English and three other Dravidian languages; 3) monolingual data for all five languages in a different domain from the test data.

Our main contributions in this paper are:
\begin{itemize}[noitemsep]
    \item We show that transliteration is beneficial in UNMT between related languages that do not share the same writing system.
    \item We investigated multiple architectures to include reference languages and enable MUNMT for diverse language pair such as English and Kannada.
    \item We analyze the influence of language similarity and show the importance of using reference languages similar to Kannada.
\end{itemize}

\section{Related Work} \label{sec:related}
Initial work on UNMT \citep{lample2018unsupervised} demonstrates the possibility of building unsupervised bilingual systems between similar and high resource languages. 
However, as shown recently \citep{guzman-etal-2019-flores, marchisio-etal-2020-unsupervised,kim-etal-2020-unsupervised}, purely unsupervised systems perform poorly in situations with low resource and distant language pairs . 
However, often it is in this very case that unsupervised translation could be helpful.
\citet{guzman-etal-2019-flores} propose a hybrid approach that translates between Nepali and English without parallel data by utilizing parallel data between Hindi and English. Specifically, the model is trained jointly with supervised loss between English and Hindi and unsupervised loss between English and Nepali.
In our work, we employ additional loss terms on top of this in order to more tightly connect the languages without parallel data.
Moreover, we analyze ways to incorporate auxiliary parallel data and investigate the correlation to language similarity.

The current state-of-the-art UNMT systems build upon multilingual pretraining \citep{liu2020mbart,garcia2020harnessing} and include a large number of languages from diverse languages families.
The high degree of multilinguality is expected to lead to better model generalization.
In this work, we zoom in on Dravidian languages and study how we can best utilize the related languages within this particular family.
We also study data scenarios with minimal auxiliary parallel data and use a training procedure that relies more on the monolingual data.
 
The languages in our setup belong to the same family, but have unique scripts. Because of this, there is little vocabulary overlap. \citet{chakravarthi-etal-2018-improving,chakravarthi-et-al:OASIcs,chakravarthi-etal-2019-wordnet,chakravarthi-etal-2019-multilingual} compare the impact of transliteration in supervised multilingual NMT and shows that it was helpful. Different from \citet{chakravarthi-etal-2019-multilingual}, the focus of our work is unsupervised translation. 
We study the role of lexical overlap in a unsupervised training pipeline, including pretraining on monolingual data and supervised training on auxiliary language pairs. 
We also highlight the information loss caused by transliteration from code-mixing in data from certain domains.

\section{Background}
\label{background}
We define $L_{P}$ as the set of language pairs with parallel data, and $L_{M}$ as the set of languages with monolingual data. 
The sample space of sentences for language $X\in L_{M}$ is $\phi_{X}$ following the terminology in \citep{li2020reference}.
Similarly, for language pair $X-Y\in L_{P}$, the sample space of parallel sentences is $\phi_{X-Y}$. 

A standard bilingual supervised model translating from language $X$ to $Y$ is trained on parallel dataset $\phi_{X-Y}$.
The model parameters $\theta$ are optimized to maximize the likelihood of target sentences in $Y$ given the source sentences in $X$:
\begin{equation} \label{eq:mt}
    \mathcal{L}_{MT}(\theta,X,Y)=\mathbf{E}_{(x,y) \sim \phi_{X-Y}}[-\log p_{\theta}(y|x)].
\end{equation}

For unsupervised translation, in absence of parallel sentences, \citet{lample2018unsupervised} propose to combine a denoising auto-encoding task with an on-the-fly back-translation task, with
the former learning language models from monolingual data, and the latter bridging the two languages.

In the auto-encoding task, the model aims to reconstruct sentences from corrupted inputs.
Specifically, a noise model $C(x)$ corrupts an input sentence $x$ by dropping and shuffling the words according to a given probability.
The loss function for auto-encoding using a monolingual dataset for language $X$ from the collection $L_{M}$ is defined as
\begin{equation} \label{eq:ae}
    \mathcal{L}_{AE}(\theta,X)=\mathbf{E}_{(x) \sim \phi_{X}}[-\log p_{\theta}(x|C(x))].    
\end{equation}

In the on-the-fly back-translation task, the model only needs  monolingual data and its loss function is defined as
\begin{equation} \label{eq:bt}
    \mathcal{L}_{BT}(\theta,X,Y)=\mathbf{E}_{(x) \sim \phi_{X}}[-\log p_{\theta}(x|\hat{y})],
\end{equation}
where $\hat{y}=\argmax_y\log p_{\theta}(y|x)$ is the intermediate translation generated in language $Y$ during training. 

The full loss function for unsupervised translation combines the auto-encoding loss in Equation \eqref{eq:ae} on $X$ and $Y$ respectively and the back-translation loss in Equation \eqref{eq:bt} between $X$ and $Y$ in both directions:
\begin{equation} \label{eq:unmt}
\begin{split}
    \mathcal{L}_{UMT}(\theta,X,Y) &= \mathcal{L}_{AE}(\theta,X) + \mathcal{L}_{AE}(\theta,Y)+\\
    &\mathcal{L}_{BT}(\theta,X,Y) + \mathcal{L}_{BT}(\theta,Y,X).
\end{split}
\end{equation}

Cross translation \citep{garcia-etal-2020-multilingual,li-etal-2020-reference} is a recently proposed technique that uses a reference language to improve the performance of unsupervised multilingual models. 
The procedure is similar to on-the-fly back-translation but additionally utilizes parallel dataset with \textit{reference languages}.
When training an unsupervised translation model from $X$ to $Y$, 
we use an additional parallel dataset $X-R$ from the collection $L_{P}$, where $R$ is the reference language.
Beside the UNMT loss, we additionally optimize the following cross translation loss function:

\begin{equation}  \label{eq:ct}
    \mathcal{L}_{CT}(\theta,X,R,Y)=\mathbf{E}_{(x-r) \sim \phi_{X,R}}[-\log p_{\theta}(r|\hat{y})],
\end{equation}

where $\hat{y}=\argmax_y\log p_{\theta}(y|x)$ is the intermediate translation.

For both on-the-fly back-translation and cross translation, the quality of intermediate translations is essential for the final UNMT model \citep{li-etal-2020-reference}.
High-quality intermediate translations further needs  similar encoder outputs for parallel sentences in the source and target language.
While the denoising auto-encoding objective is expected to encourage this, achieving this is expected to be easier between similar languages.
Therefore, in this work we focus on utilizing similar languages for UNMT.

\section{Language Similarity in Unsupervised Machine Translation}
As outlined in Section \ref{sec:related}, distant language pairs pose a great challenge to UNMT systems.
In order to apply UNMT between highly dissimilar languages, in our case English and Kannada, we explore several directions to leverage language similarity using other reference languages in the Dravidian family. 

First, we investigate the role of transliteration in UNMT, as the languages we work with do not share a common writing system.
Secondly, we evaluate multiple model architectures to enable unsupervised translation of dissimilar language pairs.
We experiment with using auxiliary data from various related reference languages in each architecture to show the influence of language similarity.
Last but not least, we propose a method to measure language similarity using $n$-gram overlap. 
Based on this metric, we show the benefit of including reference languages with high similarity score. 
\label{Methods}

\subsection{Transliteration} \label{subsec:transliteration}
Unsupervised translation is more difficult when the writings systems differ \citep{marchisio-etal-2020-unsupervised}.
Although the languages in our work belong the the Dravidian language family, they do not share a common script. 
In fact, the ones they we focus on (Kannada, Tamil, Telugu, Malayalam) all have unique scripts.
To enable a unified writing system and potentially common cross-lingual representations, we transliterate the Dravidian text into the Latin script.
As a result, when translating into Dravidian languages, the model outputs are romanized.
To restore the original script, we transliterate once again to convert the outputs back into the respective Dravidian writing system.

\begin{figure}[!ht]
\centering
\caption{Example sentence in Telugu showing information loss from transliteration and partial recovery through post-processing. Orange lines mark the words transliterated into the incorrect script as a result of code-mixing. Blue lines mark the correct forms.}
\includegraphics[width=0.45\textwidth]{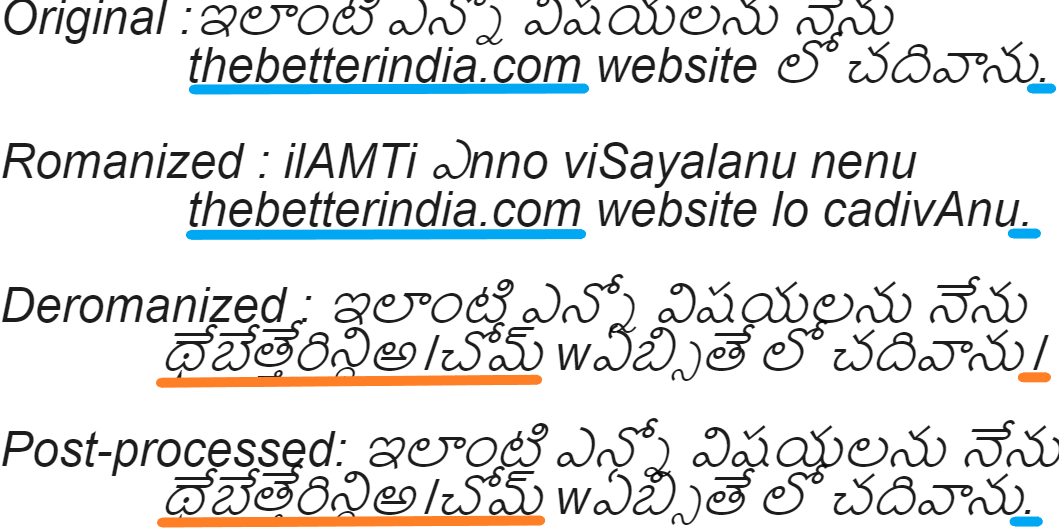}
\label{fig:transliterate}
\end{figure}

While this transliteration procedure is conceptually straightforward, in practice an additional challenge arises due to code-mixed inputs.
We illustrate this with the example in Figure \ref{fig:transliterate}.
As shown in the original Telugu sentence in the first row, the website name is written in the Latin script.
The second row shows a perfect translation in the Latin script.
However when restoring the original Telugu script in the third row, we incorrectly convert the website name into the wrong writing system, as the transliteration model has no knowledge of the code-mixing in the original sentence.
A similar issue occurs with numbers and punctuation marks, which could get converted to the Dravidian versions (e.g. ``." mapped to ``\textbar'' in Figure \ref{fig:transliterate}).

To partially cope with the problem described above, we devise a simple post-processing step.
When we restore the original script of a Dravidian language, we keep the numbers and punctuation marks in the Latin script. 
Nevertheless, we still lose information because of code-mixing in the data, such as the website name in Figure \ref{fig:transliterate}.

\subsection{Exploiting Language Similarity}
\label{multilingualUNMT}
A main focus of this work is to investigate the impact of language similarity in MUNMT. 
After allowing the model to exploit language similarity by mapping the text to a common script, we analyze the impact of various model architectures on enabling knowledge sharing. We further study the influence of language similarity in each architecture by comparing different reference languages.   

In our setup, the goal is to translate between English and Kannada but we do not have parallel data between them. Although, parallel data is available between English and reference languages such as Telugu, Tamil, Malayalam. We explain the different architectures we investigate and training procedure of these models below.

 We explore both bilingual and multilingual models to include these reference languages. We build these models by including language embedding to each word embedding as proposed in \citet{conneau2019cross}. This is similar to the inclusion of position embedding in the Transformer architecture \citet{vaswani2017attention}. As a first step, we pretrain a Masked Language Model (MLM; \citeauthor{devlin-etal-2019-bert}, \citeyear{devlin-etal-2019-bert}) on the monolingual corpora for all languages in the model following the procedure proposed by \citet{conneau2019cross}.
The goal of this step is to get better initialization of word embeddings.

\begin{figure*}[!ht]
\centering
\caption{Pictorial description of multiple architectures to include auxiliary parallel data in reference language $R$. 
Here $R$ is chosen to be a language similar to Kannada.}
\label{fig:auxiliar}
\includegraphics[width=0.9\textwidth]{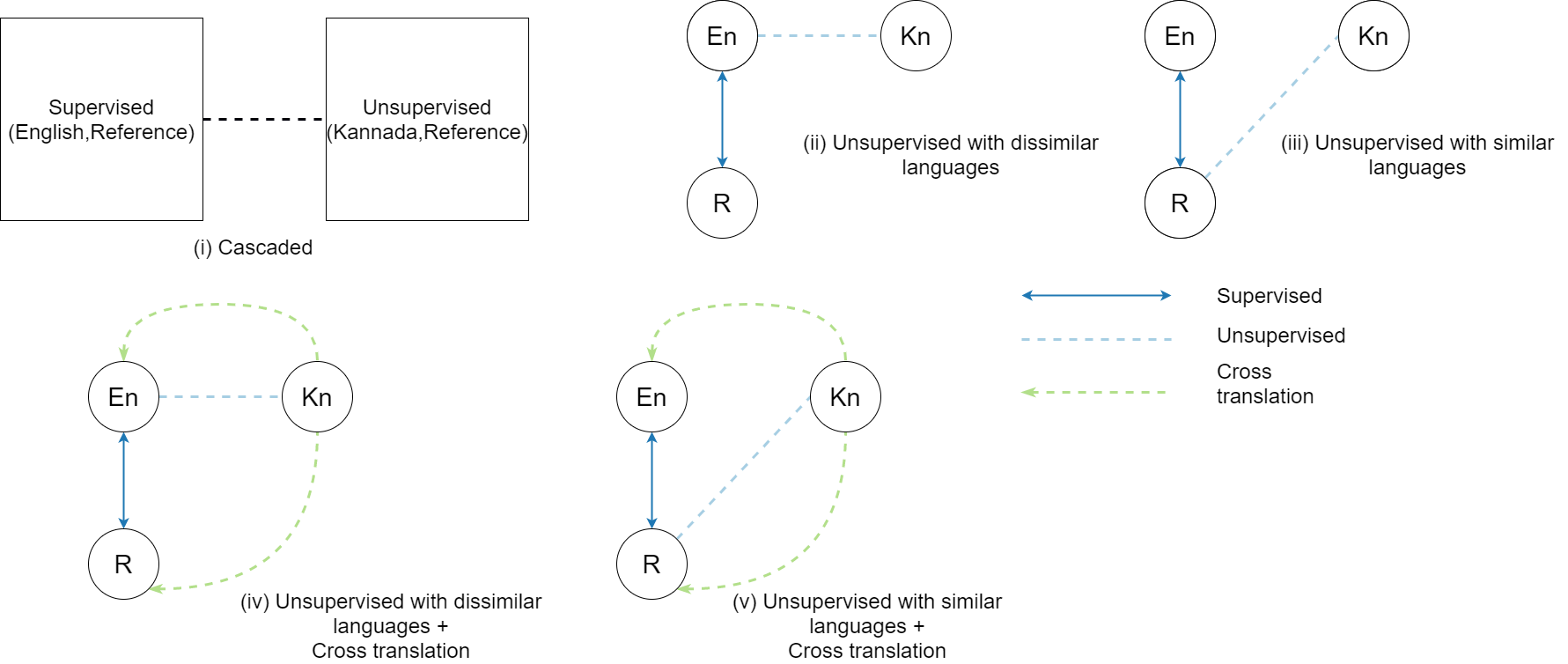}
\label{fig:config1}
\end{figure*}

After pretraining, we consider several training setups to utilize auxiliary data, as illustrated in Figure \ref{fig:auxiliar} and described next:

\paragraph{Cascaded} 
The cascaded system consists of two separate NMT models as shown in Figure 3.1, a supervised one translating English $\leftrightarrow$ reference language, and an unsupervised one translating reference language $\leftrightarrow$ Kannada.
The assumption here is that it is easier to learn unsupervised between languages that are in the same family.
The reference language will therefore be a Dravidian language.
For example, when translating from English to Kannada, we first generate a translation from English to the reference language using the supervised model and pass this intermediate translation to the unsupervised model to get the desired translation.

\paragraph{Unsupervised, dissimilar languages}  
A multilingual model is trained to jointly perform  supervised translation for reference language $\leftrightarrow$ English and unsupervised translation for English $\leftrightarrow$ Kannada as shown in Figure 3.2.
For the former, we use the standard cross-entropy loss from Equation \eqref{eq:mt}.
For the latter, we use the unsupervised loss in Equation \eqref{eq:unmt} between English and Kannada.

\paragraph{Unsupervised, similar languages} 
The setup is closely related to the previous one.
The main difference that the unsupervised translation is between Kannada and another Dravidian language, instead of English as shown in Figure 3.3.
Similar to the motivation for the cascaded system, we hope to achieve strong unsupervised model by exploiting the similarity between Kannada and the reference language. During test time, we generate translations between English and Kannada directly unlike the cascaded system, which uses intermediate translations in the reference language.

\paragraph{Unsupervised, dissimilar languages + cross translation} 
We add the cross translation loss from Equation \eqref{eq:ct} in order to generate higher-quality intermediate translations as shown in Figure 3.4.
We extend the previous setup for unsupervised between dissimilar languages but also with cross translation loss using Kannada as the language where we generate intermediate translations.

\paragraph{Unsupervised, similar languages + cross translation} 
This setup is based on the previous setup with unsupervised between similar languages as shown in Figure 3.5. We add the cross translation loss from Equation \eqref{eq:ct}, with Kannada being the language where we generate the intermediate translations.

\paragraph{Comparing the approaches}
Among the approaches outlined above, the cascaded system consists of two separately-trained NMT models, whereas all the other approaches use a single model.
Comparing to the cascaded system provides insights to the benefit of multilinguality within a single model. 
Furthermore, we will evaluate the impact of using different languages as reference language $R$.
This comparison allows us to evaluate whether a certain reference language (e.g. one similar to Kannada) is consistently more beneficial. 

\subsection{Measuring Language Similarity}
\label{overlap}
A main motivation of our approach is to utilize similar languages to improve performance on another language.
While a multitude of factors contribute to language similarity, such as morphological, syntactical, and phonetical features,
a quantitative indicator could provide us guidance when choosing auxiliary multilingual data.

One way to measure language similarity is to consider the lexical overlap between two languages. 
As the Dravidian languages we work with are agglutinative, we choose to calculate the overlap based on character $n$-grams rather than at word level. 
Since our models operate on sub-word level rather than word level, taking character $n$-grams is more appropriate. 
In addition, we take into account the frequency of the shared character $n$-grams, as overlap between rare character $n$-grams is not as important compared to frequent ones.
Below we explain the procedure for calculating the language similarity metric.

Let us consider two languages $A$ and $B$. We create their vocabularies by taking unique character $n$-grams on the monolingual data denoted as $V_A$ and $V_B$ respectively. The union between $V_A$ and $V_B$ is the list of all character $n$-grams denoted as $V$. Let $\text{freq}(\cdot)$ be a function that calculates the frequency of a character $n$-grams in the given language. 
We calculate the overlap of vocabularies between two languages A and B using the following formula
\begin{equation} \label{eq:overlap}
    \text{overlap}=\frac{\sum_{w\in V}\min(\text{freq}(A,w),\text{freq}(B,w))}{\sum_{w\in V}\max(\text{freq}(A,w),\text{freq}(B,w))},
\end{equation}
where $V = V_A \cup V_B$. Consider a scenario of two languages that do not share any common character $n$-grams. The frequency of every character $n$-grams in the union of two vocabularies is $0$ in the language that it is not present. Therefore, calculating overlap between two languages without any common character $n$-grams gives a score of $0$ as we take the minimum in the numerator. Having minimum frequency in the numerator and maximum frequency in the denominator provides higher score to character $n$-grams that occur frequently in both the languages compared to a common character $n$-grams that occurs frequently in only one language.

\section{Experiments and Results}
\label{Experiments}
First, we analyze the influence of having a unified writing system based on the approach described in Section \ref{subsec:transliteration}. 
Next, we evaluate multiple architectures with different reference languages described in Section \ref{multilingualUNMT}.
We then discuss our results in the context of language similarity.

\subsection{Datasets}
The monolingual datasets we use for Dravidian languages are from AI4Bharat-IndicNLP corpus \citep{kunchukuttan2020indicnlpcorpus} while for English we use Wikipedia dumps. 
The parallel datasets we use from English to Dravidian languages are from the PMIndia dataset \citep{haddow2020pmindia}. 
Almost $80\%$ of English sentences are common in any two parallel datasets. 
We align these common English sentences and extract parallel data between Dravidian languages. 
The number of sentences in each dataset used for training is shown in Table \ref{tab:dataset}.  

\subsection{Pre-Processing and Training Parameters}
We use the \textit{indic-transliteration}\footnote{\url{https://pypi.org/project/indic-transliteration/}} toolkit for transliteration between Latin script and Indic script. 
We then use Byte Pair Encoding \citep{sennrich-etal-2016-neural} to learn sub-words with 10k merge operations. We tokenize with \textit{Indic NLP Library}\footnote{\url{https://github.com/anoopkunchukuttan/indic_nlp_library}} for Dravidian languages whereas for English we use Moses \citep{koehn-etal-2007-moses}. 
All the experiments have been performed using \textit{XLM}\footnote{\url{https://github.com/facebookresearch/XLM}} toolkit. 
For our multilingual models, we use the Transformer \citep{vaswani2017attention} architecture with 6 layers and 8 heads whereas for our bilingual supervised model baselines we have 5 layers and 2 heads following the same configuration as in \citet{guzman-etal-2019-flores}.
We use a smaller model for the supervised model to avoid overfitting the parallel data amount is minimal. 
The other parameters are set to default values as they are in \textit{XLM}. 
The scores are reported in BLEU \citep{papineni-etal-2002-bleu}  and characTER \citep{wang-etal-2016-character}. 
When translating into Dravidian languages with transliteration, we report the scores after restoring the original scripts. 
We report detokenized BLEU scores using sacreBLEU\footnote{BLEU+case.mixed+numrefs.1+smooth.exp+tok.
    spm+version.1.4.12. We use the tokenizer ``spm'' instead of the default ``13a'' for better tokenization of Indian languages.} \citep{post-2018-call}.

\begin{table}[!ht]
\small
\centering
\caption{The number of sentences in the monolingual and parallel datasets for different languages. 
\textbf{*}Note that the English-Kannada parallel data is only used to train the supervised model (as performance upper bound).
The training of all other models do not involve this set.}
\label{tab:dataset}
\begin{tabular}{@{}lcc@{}}
\toprule
\textbf{Languages}  & 
\multicolumn{1}{l}{\begin{tabular}[c]{@{}c@{}}\textbf{Monolingual} \\ \textbf{data} \end{tabular}} &
\multicolumn{1}{l}{\begin{tabular}[c]{@{}l@{}}\textbf{Parallel data} \\ \textbf{to English} \end{tabular}} \\ 
\midrule
Kannada (Kn)   & 14M                                  & 31k\textbf{*}                                                                                       \\
Tamil (Ta)     & 21M                                  & 34k                                                                                       \\
Telugu (Te)    & 15M                                  & 34k                                                                                       \\
Malayalam (Ml) & 11M                                  & 30k                                                                                       \\
English (En)   & 46M                                  & -                                                                                         \\ \bottomrule
\end{tabular}
\end{table}

\begin{table*}[!ht]
\centering
\small
\caption{Impact of transliteration with similar and distant language pairs. Scores with noticeable differences to the other writing system are shown in bold. The translation quality is measured in BLEU ($\uparrow$) and characTER ($\downarrow$).}
\label{tab:transliteration}
\begin{tabular}{@{}cccccc|cccc@{}}
\toprule
                     &                      & \multicolumn{4}{c}{\textbf{Supervised}}                                             & \multicolumn{4}{c}{\textbf{Unsupervised}}                                          \\ \midrule
\multicolumn{1}{l}{} & \multicolumn{1}{l}{} & \multicolumn{2}{c}{Original Script}         & \multicolumn{2}{c}{Romanized}       & \multicolumn{2}{c}{Original Script}        & \multicolumn{2}{c}{Romanized}       \\
                    \midrule
Source               & Target               & BLEU  & \multicolumn{1}{l}{characTER} & \multicolumn{1}{l}{BLEU} & characTER & BLEU & \multicolumn{1}{l}{characTER} & \multicolumn{1}{l}{BLEU} & characTER \\
En                   & Kn                   & \textbf{16.4}   & 76                           & 16                     & 76       & 0.1  & 85                           & 0.2                     & 84       \\
Kn                   & En                   & \textbf{12.3}  & 77                           & 11.5                    & 77       & 0.3  & 93                           & 0.6                     & 93       \\
Te                   & Kn                   & \textbf{20.1} & 69                           & 18.9                    & 69       & 3.5 & 79                           & \textbf{7.1}                     & \textbf{74}       \\
Kn                   & Te                   & \textbf{11.9}  & 74                           & 11.1                     & 74       & 2.3 & 86                           & \textbf{5.1}                     & \textbf{82}       \\ \bottomrule
\end{tabular}
\end{table*}
\subsection{Effect of Unifying Writing System}
In order to assess the impact of transliteration, we perform experiments in the supervised and unsupervised setup between the following language pairs:
\begin{itemize}[noitemsep] 
    \item English and a Dravidian language (dissimilar).
    \item Two Dravidian languages (similar).
\end{itemize}

We consider English and Kannada for dissimilar language pair and Kannada and Telugu for similar pair.
In Table \ref{tab:transliteration}, we report the performance before and after transliteration.

First, for the supervised models, romanizing the Dravidian languages has no conclusive effects on translation quality. 
While there is a drop in BLEU points after transliteration, the characTER scores remain unchanged.
Upon inspection of the translation results, we also notice some of the outputs suffer from the information loss described in Section \ref{subsec:transliteration}.

Second, we observe that the purely unsupervised systems perform poorly, especially between English and Kannada, the dissimilar language pair. 
The scores are nearly zero with or without romanization. 
This confirms the previous findings \cite{marchisio-etal-2020-unsupervised,kim-etal-2020-unsupervised} that unsupervised translation is difficult between distant languages.
Between Telugu and Kannada, the unsupervised performance largely improves after romanization.
The BLEU scores increase by $3.6$ and $2.8$ points respectively when translating into and from Kannada. The characTER scores also decrease by 5 and 4 points absolute.
This demonstrates that mapping the text into a common script helps in exploiting the similarity between related languages for UNMT.
This finding also contrasts the observations under the supervised model, where transliteration shows no conclusive impact.
The difference leads use to conclude that the role of transliteration is scenario-dependent, in this case on the amount of available supervision signals from parallel data.

Furthermore, we use the similarity metric proposed in Section \ref{overlap} to validate this finding. We calculate this by taking character tri-grams on the monolingual data.
Indeed, in Table \ref{tab:overlap}, we see that the overlap between Kannada and Telugu increase from $1\%$ to $34\%$ after transliteration, exhibiting the highest similarity scores between Kannada and \{Te, Ta, Ml\}.
This finding is also in line with the close relation between Kannada and Telugu from the linguistic literature \cite{datta1988encyclopaedia,bright1996kannada}.
This evidence shows that transliteration bridges across different writing systems and enables similar representations for related languages.
After transliteration, the unsupervised model can more easily leverage the similarity between these related languages and therefore achieve better performance.

\begin{table}[h]
\centering
\small
\caption{Lexical overlap (\%; calculated using Equation \ref{eq:overlap}) before and after transliteration (before $\rightarrow$ after).}
\label{tab:overlap}
\begin{tabular}{@{}llllcc@{}}
\toprule
   & En                       & Kn                       & Te                       & \multicolumn{1}{l}{Ta}                       & \multicolumn{1}{l}{Ml} \\ \midrule
En & \multicolumn{1}{c}{100}    & \multicolumn{1}{c}{0 $\rightarrow$ 6}    & \multicolumn{1}{c}{0 $\rightarrow$ 6}    & 0 $\rightarrow$ 3                                            & 0 $\rightarrow$ 6                      \\
Kn & \cellcolor[HTML]{9B9B9B} & \multicolumn{1}{c}{100}    & \multicolumn{1}{c}{1 $\rightarrow$ \textbf{34}}   & 1 $\rightarrow$ 10                                           & 1 $\rightarrow$ 26                     \\
Te & \cellcolor[HTML]{9B9B9B} & \cellcolor[HTML]{9B9B9B} & \multicolumn{1}{c}{100}    & 1 $\rightarrow$ 9                                            & 1 $\rightarrow$ 27                     \\
Ta & \cellcolor[HTML]{9B9B9B} & \cellcolor[HTML]{9B9B9B} & \cellcolor[HTML]{9B9B9B} & 100                                            & 1 $\rightarrow$ 11                     \\
Ml & \cellcolor[HTML]{9B9B9B} & \cellcolor[HTML]{9B9B9B} & \cellcolor[HTML]{9B9B9B} & \multicolumn{1}{l}{\cellcolor[HTML]{9B9B9B}} & 100                      \\ \bottomrule
\end{tabular}
\end{table}

\subsection{Impact of Model Architectures}
\label{experimentsreference}
We evaluate the setups described in Section \ref{multilingualUNMT} with different reference languages and report these scores in Table \ref{tab:auxiliary}. 
 Given the previously-found positive impact of unifying the writing systems on unsupervised learning, we transliterate Dravidian languages in all experiment setups here.

First, the unsupervised models on similar languages consistently outperform those with dissimilar languages. 
This is in line with previous results in Table \ref{tab:transliteration} and further validates our finding on the positive role of language similarity in UNMT.

Second, adding the cross translation loss improves the translation quality into English.
The model using Telugu as reference language achieves $5.1$ BLEU points when translating into English, the highest among the setups explored so far.
However, there is not a similar gain when translating into Kannada.
One reason is that cross translation generates more training signals when translating into English but 
does not explicitly encourage better translation into Kannada.

\begin{table*}[]
\centering
\caption{Unsupervised translation performance between Kannada and English using different reference languages in multiple setups described in Section \ref{multilingualUNMT}. Scores are shown as x/y where x indicates in Kn $\rightarrow$ En direction and y indicates in En $\rightarrow$ Kn respectively. 
The translation quality is measured in BLEU ($\uparrow$) and characTER ($\downarrow$). 
The best scores (BLEU \& characTER) for each reference language in both directions are shown in bold.}

\label{tab:auxiliary}
\resizebox{\linewidth}{!}{%
\begin{tabular}{@{}ccccccccccc@{}}
\toprule
                                                              & \multicolumn{2}{c}{Cascaded} & \multicolumn{2}{c}{Unsupervised dissimilar} & \multicolumn{2}{c}{Unsupervised similar} & \multicolumn{2}{c}{\begin{tabular}[c]{@{}c@{}}Unsupervised dissimilar\\ + Cross translation\end{tabular}} & \multicolumn{2}{c}{\begin{tabular}[c]{@{}c@{}}Unsupervised similar\\ + Cross translation\end{tabular}} \\ \midrule
\begin{tabular}[c]{@{}c@{}}Reference \\ Language\end{tabular} & BLEU          & characTER(\%)     & BLEU                  & characTER(\%)            & BLEU                & characTER(\%)           & BLEU                                              & characTER(\%)                                              & BLEU                                               & characTER(\%)                                          \\
\midrule
Te                                                            & 3.7/\textbf{1.6}     & 86/\textbf{82}        & 1.3/2.1             & 89/87               & 3.8/1           & 86/85              &2.2/0.4                                                   &87/90                                                       & \textbf{5.1}/1.4                                           & \textbf{85}/93                                             \\
Ml                                                            & 2.3/\textbf{1.1}     & 86/\textbf{87}        & 1.1/0.6              & 89/92               & 2.0/0.7            & 89/93              &2.7/0.9                                                   &85/91                                                       & \textbf{3.3}/0.2                                           & \textbf{84}/96                                             \\
Ta                                                            & 1.2/0.2     & 87/\textbf{89}        &0.3/\textbf{0.4}                       &89/93                     & 1.4/0.4           & 88/94            &0.7/0.3                                                   &\textbf{86}/94                                                       & \textbf{1.8}/0.1                                            & 87/97                                             \\ \bottomrule
\end{tabular}%
}
\end{table*}

\subsection{Impact of Reference Languages}
Table \ref{tab:auxiliary} provides a clear view of which reference language is more beneficial to a purely unsupervised system.
In line with previous observations, when using Telugu as the reference language, the model performance exceeds when using other languages. 
The only exception is unsupervised dissimilar + cross translation scenario where it falls slightly behind Malayalam.
The positive impact of adding Telugu is within expectation given the previous finding on its close relation to Kannada.
After Telugu, the next best choices for the reference languages are Malayalam, followed up Tamil. 
This particular order agrees with the vocabulary overlap with Kannada shown in Table \ref{tab:overlap}, with the highest value on Telugu ($34\%$) followed by Malayalam ($26\%$) and lastly Tamil ($10\%$). 
Based on our similarity metric, we are able to show the impact of using languages that are more similar. 

Given the positive impact of incorporating related reference languages, we extend our investigation to using multiple reference languages.
We build upon the best architecture (unsupervised similar + cross translation) and train it with using auxiliary data of two reference languages. 
We choose Telugu and Malayalam as they are more related to Kannada and yielded stronger performance than Tamil.
The scores for this along with our baselines are in Table \ref{tab:multiple}. 
By additionally incorporating Malayalam, we see an improvement from $5.1$ to $5.4$ BLEU points when translating to English. The overlap in the English sentences for the parallel data with the reference languages is more than $80\%$. Even though we are adding additional data, very few new in-domain English sentences are seen by the model. This might explain the minimal increase in score after adding multiple reference languages. 
While this performance is still $6.1$ points short from supervised model, it is significantly better than the purely unsupervised baseline when translating into English.

\begin{table}[]
\centering
\caption{ Scores for baselines and proposed model that includes reference languages. MUNMT corresponds to unsupervised similar + cross translation architecture. 
``MUNMT, Te'' uses only Telugu as a reference language whereas ``MUNMT, Te$+$Ml" uses both Telugu and Malayalam as reference languages.}
\label{tab:multiple}
\resizebox{\columnwidth}{!}{%
\begin{tabular}{@{}lcccc@{}}
\toprule
             & \multicolumn{2}{c}{Kn $\rightarrow$ En}                               & \multicolumn{2}{c}{En$\rightarrow$ Kn} \\ \midrule
             & \multicolumn{1}{l}{BLEU} & \multicolumn{1}{l}{characTER(\%)} & BLEU      & characTER(\%)      \\
             \midrule
Supervised   & 11.5                    & 77                           & 16      & 76            \\
Unsupervised & 0.6                     & 93                           & 0.2      & 84            \\
MUNMT, Te 
    & 5.1                     & 85                           & 1.4      & 93            \\
MUNMT, Te$+$Ml  & \textbf{5.4}                      & \textbf{84}                           & 1.6      & 95            \\ \bottomrule
\end{tabular}%
}
\end{table}

\subsection{Examples of Kn$\rightarrow$En Translation}
\begin{figure} 
    \centering
    \includegraphics[width=0.45\textwidth]{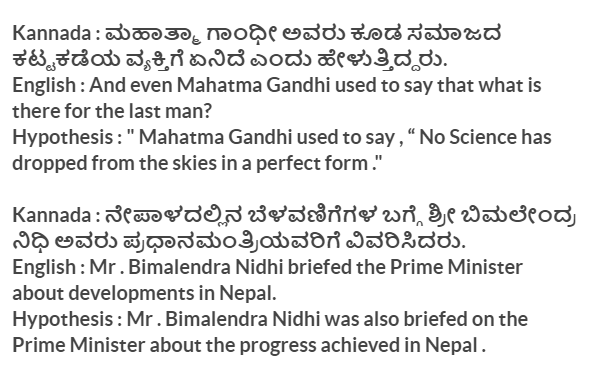}
    \caption{Example outputs when translating Kn$\rightarrow$En.}
    \label{fig:example_kn-en}
\end{figure}

We highlight that our overall task for unsupervised translation remains very challenging, as the auxiliary parallel data amount with the reference languages is also minimal ($\sim$30k sentences).
Moreover, there is a domain discrepancy between the monolingual data and the test data from government announcements.
In Figure \ref{fig:example_kn-en}, we show example outputs when translating Kn$\rightarrow$En using the previous best ``MUNMT, Te+Ml" model.
The first example shows the problem of hallucination \cite{koehn-knowles-2017-six}, hinting towards negative impacts of domain shifts.
Nonetheless, the model sometimes achieves translation with adequate quality such as the last example in Figure \ref{fig:example_kn-en}.

\section{Conclusion}
We focus on the challenging yet realistic scenario of translating between English and Kannada without any parallel data.
After confirming the difficulty of purely unsupervised translation between distant language pairs, we show that increasing the lexical overlap by transliteration is necessary for achieving UNMT for our focal language.
Moreover, we utilize monolingual and limited amount of auxiliary parallel data in languages related to Kannada.
The best performance is achieved by UNMT between similar languages along with cross translation.
We show the importance of incorporating languages with high relatedness to the focus language by evaluating different reference languages.
Finally, we propose a metric for measuring lexical overlap
and show that it serves as a good indicator for selecting the best reference languages.

\paragraph{Acknowledgement}
This work is supported by the Facebook Sponsored Research Agreement  ``Language Similarity in Machine Translation".
\bibliographystyle{acl_natbib}
\bibliography{main.bib,anthology.bib}
\end{document}